\documentclass{article}

    \usepackage[preprint, nonatbib]{neurips_2020}

\usepackage[utf8]{inputenc} 
\usepackage[T1]{fontenc}    
\usepackage{hyperref}       
\usepackage{url}            
\usepackage{booktabs}       
\usepackage{amsfonts}       
\usepackage{nicefrac}       
\usepackage{microtype}      

\title{Survey on Self-supervised Representation Learning Using Image Transformations}

\author{%
  Muhammad Ali${}^{*}$ \qquad Sayed Hashim\thanks{First authors}\\
  Mohamed Bin Zayed University of Artificial Intelligence, UAE \\
  \texttt{\{muhammad.ali, sayed.hashim\}@mbzuai.ac.ae} \\
    \\
}

\begin{document}
\maketitle

\begin{abstract}
Deep neural networks need huge amount of training data, while in real world there is a scarcity of data available for training purposes. To resolve these issues, self-supervised learning (SSL) methods are used. SSL using geometric transformations (GT) is a simple yet powerful technique used in unsupervised representation learning. 
Although multiple survey papers have reviewed SSL techniques, there is none that only focuses on those that use geometric transformations. Furthermore, such methods have not been covered in depth in papers where they are reviewed. Our motivation to present this work is that geometric transformations have shown to be powerful supervisory signals in unsupervised representation learning. Moreover, many such works have found tremendous success, but have not gained much attention.
We present a concise survey of SSL approaches that use geometric transformations. We shortlist six representative models that use image transformations including those based on predicting and autoencoding transformations. We review their architecture as well as learning methodologies. We also compare the performance of these models in the object recognition task on CIFAR-10 \cite{cifar} and ImageNet \cite{imagenet} datasets. 
Our analysis indicates the AETv2 \cite{aetv2} performs the best in most settings. Rotation with feature decoupling \cite{decoupling} also performed well in some settings. We then derive insights from the observed results. Finally, we conclude with a summary of the results and insights as well as highlighting open problems to be addressed and indicating various future directions. 

\textbf{Keywords}: deep learning, self-supervised learning, geometric transformations, autoencoding, rotation, representation learning

\end{abstract}

\section{Introduction}
\subsection{Background}
Deep neural networks, especially convolutional neural networks (CNNs) has led to big accomplishments in the computer vision field \cite{rotation}. For real-world problems,  there is a scarcity of data for training purposes, so expensive efforts concerning time and resources are required to provide these labelled training data.  This problem led to a big increase in the interest of researchers in using unsupervised feature learning for solving visual understanding tasks with lack of availability of  labelled data \cite{aet}. 
The basic idea in SSL is to produce some supervisory signals to solve assigned tasks. This task may include representation of the data or auto labelling of the data. SSL provides us with the advantage of training networks without requiring extensive labelling of the images. In order to utilize the strengths of SSL techniques over the period of time, major area of work has been done in the domain of developing various pretext tasks.
\subsection{Previous studies}

Multiple survey and review papers on self-supervised techniques have been published. One such work \cite{surveydeep} describes deep learning-based self-supervised general visual feature learning methods from images or videos.A survey on augmentation techniques was published in late 2019 but it majorly focused on application of augmentation techniques including geometric transformation, color space augmentations, filters and feature space augmentations. Another paper \cite{surveygencon}  explained self-supervised learning technique in various fields including computer vision,graph learning and natural language processing based on  generative, contrastive and mix of generative and contrastive objectives . Another paper \cite{surveycontrast}  discusses the use of contrastive learning approach with self supervised methods.Basic idea in this technique is to assemble similar samples near each other in relevant to other samples with dissimilarities.

\subsection{Problem statement and motivation}

To the best of our ability, we were unable to find a review or survey paper that solely focuses on self-supervised learning (SSL) techniques that use geometric transformations (GT). Although many such works have been reviewed in review papers, these have not been covered in depth. For instance, one of the review papers \cite{surveydeep} talks about the popular rotation prediction SSL technique, but doesn't go into the details of it and doesn't mention works which have improvised on the method. Also, it doesn't talk about the much successful work of autoencoding geometric transformations \cite{aet}. This lack of extensive review of geometric transformations based SSL techniques show  a gap in the literature which we aim to fill.

We believe an extensive review that purely focuses on geometric transformation based self-supervised techniques is necessary due to 2 reasons. (1) Geometric transformations have proved to be simple yet powerful supervisory signals in unsupervised representation learning. (2) Many successful works have used geometric transformations from different paradigms such as autoencoding and classification. To enable a detailed and in-depth understanding of SSL with geometric transformations we present a concise survey encompassing multiple aspects including algorithmic details as well as futuristic unresolved problems.

\subsection{Proposed approach and contributions}
This work contains thorough review and comparison of 6 SSL models that use GT. Multiple avenues including preprint servers, conference proceedings and journals were searched for shortlisting relevant papers. Our contributions can be summarized in 4 points. (1) Provide the reader with an in-depth review of geometric transformation based SSL techniques. (2) Highlight the importance and success of using such methods. (3) Discuss the shortcomings in such techniques and the relevant problems. (4) Explain trend setting research as well as future direction.

We organize the rest of the paper as follows. In section 3 we briefly discuss overview of existing methods in detail, where we describe methods that are based on image rotation prediction as well as using auto encoding transformations. In section 4 we present results along with qualitative and quantitative comparison of the performance of the techniques on CIFAR-10 \cite{cifar} and ImageNet \cite{imagenet} datasets. In section 5 we derive insights and explain the results of comparing the different approaches. Section 6 contains conclusion and future directions.

\section{Overview of existing methods}
\subsection{Methods based on predicting geometric transformations}
\begin{figure}[h]
    
    \caption{Depiction of the rotation feature decoupling method. The ouput from neural network is a decoupled meaningful feature representation comprising sections that are related and unrelated to rotation. The rotation related part is trained by predicting rotations applied to image. A PU learning problem is modelled here with using the noisy rotation labels, that is trained to learn instance weights in order to decrease the impact of pictures that are rotation ambiguous. The second part is trained using loss that penalises distance in order to impose rotation irrelevance along with a task of discriminating instances by the use of classification without parameters}
    
    \includegraphics[width=13cm]{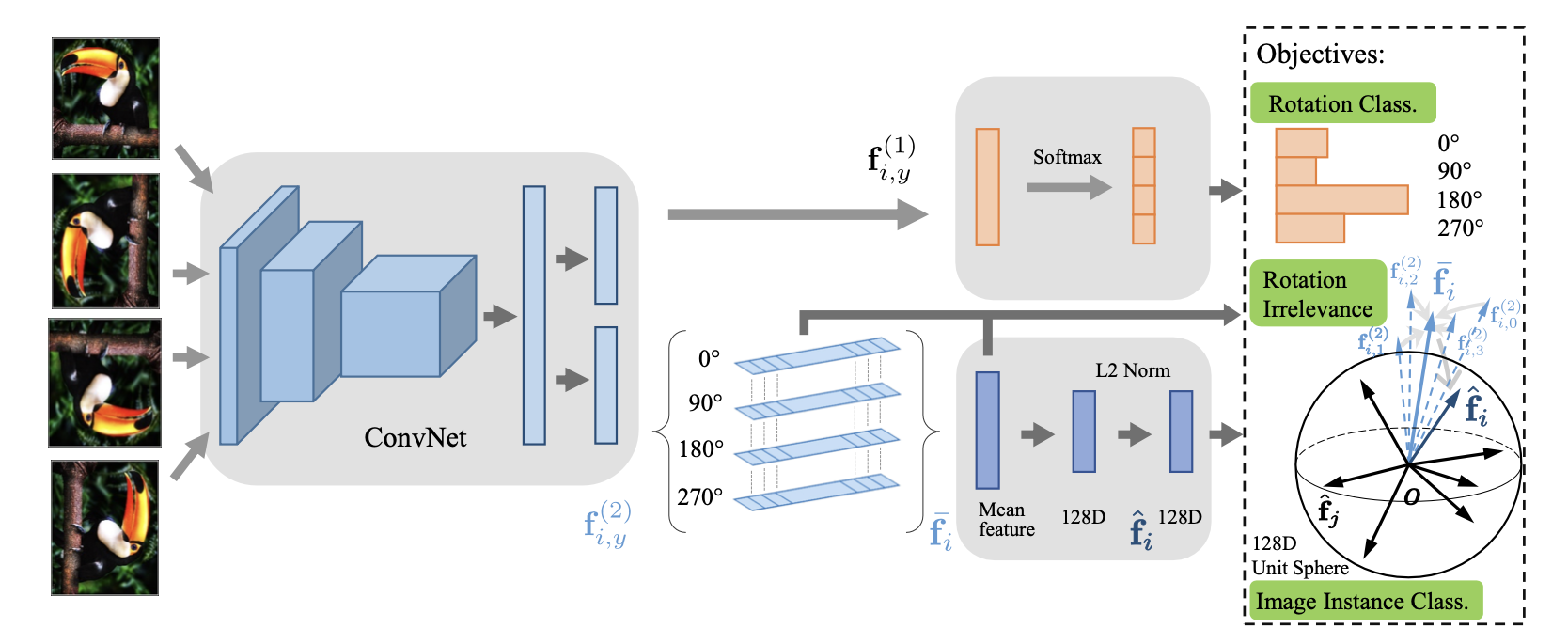} 
    \centering
    \label{fig:decoupling}
\end{figure}

In order to extract useful features from images in unsupervised fashion, RotNet \cite{rotation} is trained to predict the rotation by multiples of 90 degrees applied to the input images. 
Features learned through this method are capable of generalizing well in various tasks. But the features learned are biased with regards to rotation transformation and does not help in various tasks which are in favour of invariance in rotation. Also, rotation is not determinable for all images in practice and therefore this method does not perform well with orientation agnostic images. To address these shortcomings, an SSL technique \cite{decoupling} that decouples the learned features using the task of predicting rotation along with the task of task discriminating the instances was presented. The learned representations comprises of 2 parts in which one is discriminative to rotation and other is unrelated to rotation as shown in Figure \ref{fig:decoupling}. In another work called ExemplarCNN \cite{exemplarcnn}, authors trained the CNN to differentiate among a group of surrogate classes. To create such classes, various transformations are applied picture patch sampled randomly. This  algorithm strikingly performed well.

    
    
\subsection{Methods based on autoencoding geometric transformations}
For a randomly sampled transformation, the self-supervised learning technique of Auto-Encoding Transformation (AET) \cite{aet} attempts to quantify it just using the learned features precisely as the output. The main theory here is that the transformation can be quantified if the learned features are able store the important characteristics of original images and images after transformation. A variant of this approach is Autoencoding Variational Transformations (AVT) \cite{avt}. Provided images after transformation, AVT trains the autoencoder by increasing the shared features between learned representations and transformations. 

In order to quantify the divergence of predicted transformations from their labelled equivalents, deterministic and probabilistic AETs depend upon Euclidean distance. But, this is a contentious assumption as a set of transformations usually stay on a curved manifold instead of residing in a flat Euclidean space. To solve this issue, the authors developed AETv2 \cite{aetv2}, which uses the geodesic distance to distinguish the way in which an image is transformed towards the manifold of a set of transformations, and use its length to quantify the difference between transformations. 


\begin{figure}[h]
    
    \caption{Depiction of the way in which AETv2 \cite{aetv2} is trained from start to finish. In order to make the output matrix $\mathbf{T}^{-1} \hat{\mathbf{T}}$ from the decoder of transformation possess a unit determinant, it is normalized, and it should be noted that the projection of $\mathbf{T}^{-1} \hat{\mathbf{T}}$ onto $\mathbf{S O}(3)$ is followed in order to calculate the geodesic distance and also the loss of projection in order for the model to be trained}
    
    \includegraphics[height=7cm]{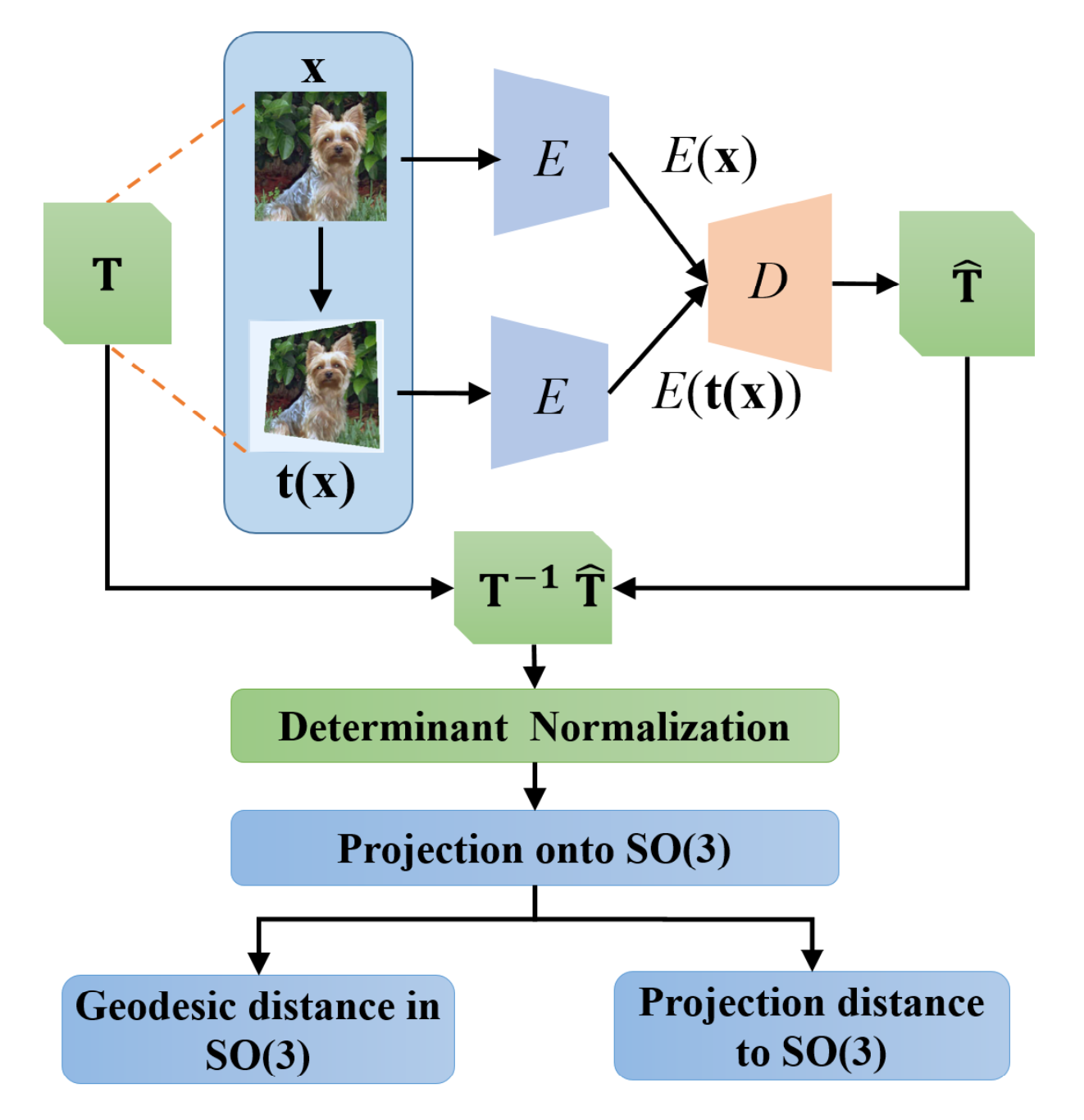} 
    \centering
    \label{fig:aetv2}
\end{figure}

\section{Results}
\subsection{CIFAR-10}

Comparison of the performance of the models in the downstream task of object recognition on CIFAR-10 \cite{cifar} measured by error rates produced interesting results, as shown in Table \ref{tab:cifar}. AETv2 \cite{aetv2} model produced best results with an error rate of 7.44. It can also be seen that RotNet \cite{rotation} was a breakthrough work, that made much progress compared to previous traditional methods, bringing down the error rate to 8.84. While rotation provides a simple, yet powerful supervisory signal, autoencoding multiple transformations has proved to be a slightly better method in feature representation learning.
\begin{table}[h]
\caption{ Comparative study between various selected methods using self supervised technique  on CIFAR-10 \cite{cifar} data set.Two supervised methods including NIN and random Init along with conv contains same architecture with only difference that  one  is supervised where as the other one is trained in such a way that first two blocks are initialized and kept frozen in training process.}
\centering
\begin{tabular}{cc}
\hline
Model                             & Error rate    \\ \hline
NIN-Supervised (Lower Bound)      & 7.20          \\
Random Init. + conv (Upper Bound) & 27.50         \\ \hline
Roto-Scat + SVM  \cite{rotoscat}                 & 17.7          \\
DCGAN  \cite{dcgan}                 & 17.2          \\
ExamplarCNN \cite{exemplarcnn}              & 15.7          \\
Scattering \cite{scattering}  & 15.3          \\
RotNet + conv  \cite{rotation}   & 8.84          \\
AETv1 + conv \cite{aet}   & 7.82          \\
AVT + conv   \cite{avt}    & 7.75 \\
AETv2 + conv \cite{aetv2}    & \textbf{7.44} \\ \hline
\end{tabular}
\label{tab:cifar}
\end{table}

\subsection{ImageNet}

Some interesting insights could be derived from the results of the methods' performance in the image classification task on ImageNet \cite{imagenet} compared using top-1 accuracy, as presented in Table \ref{tab:imagenet}. While AETv2 performed better when the network was trained with up to 3 convolutional blocks, RotNet performed better with more than 3 convolutional blocks. This could be due to the simplicity of the RotNet model. Generally, using more than 3 convolutional blocks produced a gradual decrease in object recognition accuracy, which we believe is because the feature learnt in these layers start to become more specific on the pretext task. Furthermore, we see that increased depth of models led to better performance in object recognition with regards to the feature maps produced by earlier layers. We believe this is because a deeper model enables the features of layers early on to be less peculiar to the pretext task.

\begin{table}[h]
\caption{Comparison of Top-1 accuracy using linear layers on  ImageNet \cite{imagenet} data set. For comparison of self supervised models Alex Net is used. This classifier is trained with various depth of convolution layers containing features maps with 9000 elements finally.Upper bound and lower bounds of self supervised model's performances for  supervised and random models are also shown for comparison.}
\centering
\begin{tabular}{cccccc}
\hline
Model                         & Conv1         & Conv2         & Conv3         & Conv4         & Conv5         \\ \hline
ImageNet Labels (Upper Bound) \cite{rotation} & 19.3          & 36.3          & 44.2          & 48.3          & 50.5          \\
Random (Lower Bound) \cite{rotation}         & 11.6          & 17.1          & 16.9          & 16.3          & 14.1          \\
Random rescaled \cite{randomrescaled} (Lower Bound) & 17.5          & 23.0          & 24.5          & 23.2          & 20.6          \\ \hline
BiGAN  \cite{bigan}                      & 17.7          & 24.5          & 31.0          & 29.9          & 28.0          \\
RotNet  \cite{rotation}                      & 18.8          & 31.7          & 38.7          & 38.2          & 36.5          \\
Rotation feature decoupling \cite{decoupling}  & 19.3          & 33.3          & 40.8          & \textbf{41.8} & \textbf{44.3} \\
AETv1  \cite{aet}                       & 19.2          & 32.8          & 40.6          & 39.7          & 37.7          \\
AVT    \cite{avt}                       & 19.5          & 33.6          & 41.3          & 40.3          & 39.1          \\
AETv2  \cite{aetv2}                       & \textbf{19.6} & \textbf{34.1} & \textbf{41.9} & 40.4          & 37.9       \\ \hline  
\end{tabular}
\label{tab:imagenet}
\end{table}

\section{Discussion}
\begin{figure}
     \centering
     \begin{subfigure}[b]{0.48\textwidth}
         \centering
         \includegraphics[width=\textwidth]{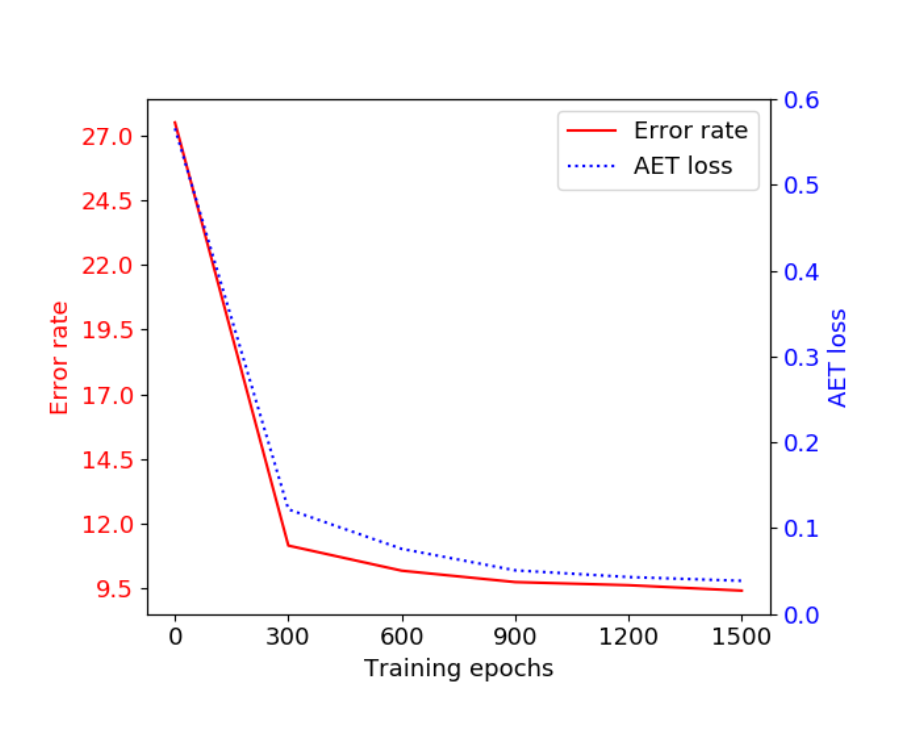}
         \caption{CIFAR-10 \cite{imagenet}}
         \label{fig:aetcifar}
     \end{subfigure}
     \hfill
     \begin{subfigure}[b]{0.48\textwidth}
         \centering
         \includegraphics[width=\textwidth]{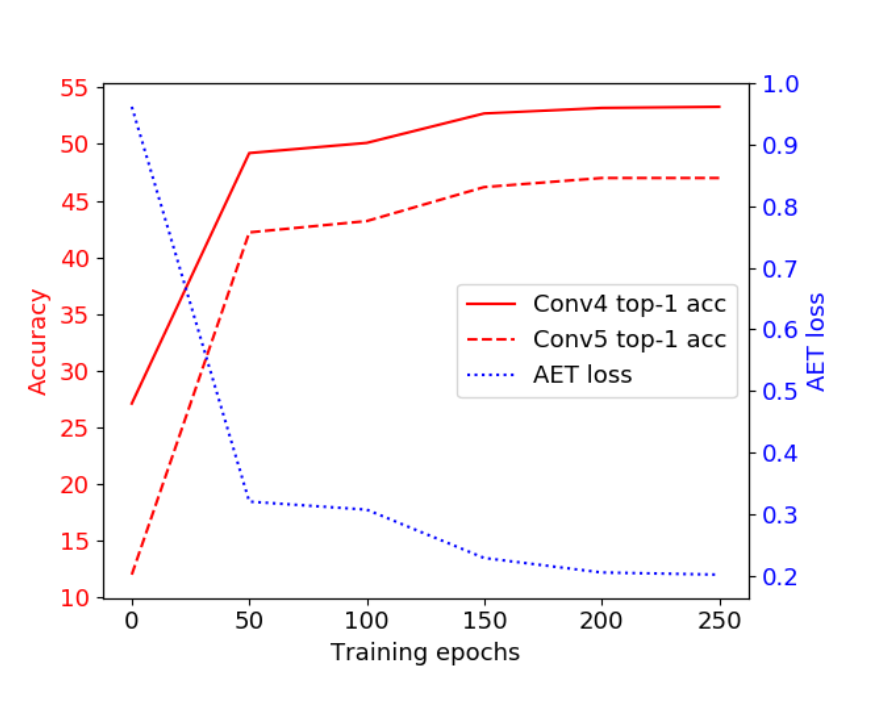}
         \caption{ImageNet \cite{imagenet}}
         \label{fig:aetimagenet}
     \end{subfigure}
     \hfill
     
    \caption{The two plots show the change in error rate and top-1 accuracy with the change in AET \cite{aet} loss as training progressed on the CIFAR-10 \cite{imagenet} and ImageNet \cite{imagenet} datasets.}
    \label{fig:aetloss}
\end{figure}

A thorough comparison of the methods have shown that AETv2 \cite{aetv2} performs best in terms of learning relevant features from images, and being able to contribute to lower error rates in object recognition on ImageNet \cite{imagenet} and CIFAR-10 \cite{cifar} datasets. As shown in Figure \ref{fig:aetloss}, it can be seen that the path of loss of prediction of transformation follows a similar path of that of error of classification and top-1 accuracy on CIFAR-10 \cite{cifar} and ImageNet \cite{imagenet}. This indicates that predicting transformations better implies better result of classification making use of learned representations. This validates the choice of AET and its variants for the supervision of learning feature representations.

One downside of using transformations is that they might leave behind any low-level visual artifacts that are easily detectable which will lead the CNN to learn easy features without practical value in the tasks of vision perception, e.g., to implement scale and aspect ratio image transformations, image resizing routines that leave easily detectable image artifacts would have to be used. The models' performance on the CIFAR-10 \cite{cifar} and ImageNet \cite{imagenet} datasets can still be improved. More combinations of transformations could be experimented with to improve performance. Moreover, attention based models could be coupled with the transformations to facilitate better feature learning.

\section{Conclusion}
Although multiple papers have reviewed contrastive and generative SSL techniques there is none that solely focuses on SSL techniques that use GT. Also, such methods have not been covered in depth in papers where they have been reviewed. This is a gap in the literature which we aim to fill through this work. Our motivation to present this work is that GT have shown to be powerful supervisory signals in unsupervised representation learning. Moreover, many works which used GT in SSL techniques have found tremendous success, but have not gained much attention. The work has extensively reviewed 6 approaches on self-supervised representation learning using image transformations that are based on predicting rotation as well as auto encoding transformations. We compared the models based on their performance in the downstream task of object recognition on CIFAR-10 \cite{cifar} and ImageNet \cite{imagenet} datasets, measured by their error rates and top-1 accuracy. The review provided insights as to which models worked better and why. It has also showed that image transformations are powerful supervisory signals for feature representation learning. We also believe this paper will help researchers in inventing new GT based SSL techniques.

\clearpage
\bibliographystyle{abbrv}
\bibliography{sample.bib}

\begin{thebibliography}{10}

\bibitem{imagenet}
J.~Deng, W.~Dong, R.~Socher, L.-J. Li, K.~Li, and L.~Fei-Fei.
\newblock Imagenet: A large-scale hierarchical image database.
\newblock In {\em 2009 IEEE conference on computer vision and pattern
  recognition}, pages 248--255. Ieee, 2009.

\bibitem{bigan}
J.~Donahue, P.~Kr{\"{a}}henb{\"{u}}hl, and T.~Darrell.
\newblock Adversarial feature learning.
\newblock {\em CoRR}, abs/1605.09782, 2016.

\bibitem{exemplarcnn}
A.~Dosovitskiy, J.~T. Springenberg, M.~Riedmiller, and T.~Brox.
\newblock Discriminative unsupervised feature learning with convolutional
  neural networks.
\newblock In Z.~Ghahramani, M.~Welling, C.~Cortes, N.~Lawrence, and K.~Q.
  Weinberger, editors, {\em Advances in Neural Information Processing Systems},
  volume~27. Curran Associates, Inc., 2014.

\bibitem{decoupling}
Z.~Feng, C.~Xu, and D.~Tao.
\newblock Self-supervised representation learning by rotation feature
  decoupling.
\newblock {\em 2019 IEEE/CVF Conference on Computer Vision and Pattern
  Recognition (CVPR)}, 2019.

\bibitem{rotation}
S.~Gidaris, P.~Singh, and N.~Komodakis.
\newblock Unsupervised representation learning by predicting image rotations,
  2021.

\bibitem{surveycontrast}
A.~Jaiswal, A.~R. Babu, M.~Z. Zadeh, D.~Banerjee, and F.~Makedon.
\newblock A survey on contrastive self-supervised learning, 2021.

\bibitem{surveydeep}
L.~Jing and Y.~Tian.
\newblock Self-supervised visual feature learning with deep neural networks:
  {A} survey.
\newblock {\em CoRR}, abs/1902.06162, 2019.

\bibitem{cifar}
A.~Krizhevsky.
\newblock Learning multiple layers of features from tiny images.
\newblock {\em University of Toronto}, 05 2012.

\bibitem{randomrescaled}
P.~Krähenbühl, C.~Doersch, J.~Donahue, and T.~Darrell.
\newblock Data-dependent initializations of convolutional neural networks,
  2016.

\bibitem{aetv2}
F.~Lin, H.~Xu, H.~Li, H.~Xiong, and G.-J. Qi.
\newblock Aetv2: Autoencoding transformations for self-supervised
  representation learning by minimizing geodesic distances in lie groups, 2019.

\bibitem{surveygencon}
X.~Liu, F.~Zhang, Z.~Hou, Z.~Wang, L.~Mian, J.~Zhang, and J.~Tang.
\newblock Self-supervised learning: Generative or contrastive, 2021.

\bibitem{scattering}
E.~{Oyallon}, E.~{Belilovsky}, and S.~{Zagoruyko}.
\newblock Scaling the scattering transform: Deep hybrid networks.
\newblock In {\em 2017 IEEE International Conference on Computer Vision
  (ICCV)}, pages 5619--5628, 2017.

\bibitem{rotoscat}
E.~Oyallon and S.~Mallat.
\newblock Deep roto-translation scattering for object classification.
\newblock {\em CoRR}, abs/1412.8659, 2014.

\bibitem{avt}
G.-J. Qi, L.~Zhang, C.~W. Chen, and Q.~Tian.
\newblock Avt: Unsupervised learning of transformation equivariant
  representations by autoencoding variational transformations, 2019.

\bibitem{dcgan}
A.~Radford, L.~Metz, and S.~Chintala.
\newblock Unsupervised representation learning with deep convolutional
  generative adversarial networks, 2016.

\bibitem{aet}
L.~Zhang, G.-J. Qi, L.~Wang, and J.~Luo.
\newblock Aet vs. aed: Unsupervised representation learning by auto-encoding
  transformations rather than data, 2021.

\end{thebibliography}








\end{document}